\documentclass[conference]{IEEEtran}
\IEEEoverridecommandlockouts
\usepackage{cite}
\usepackage{amsmath,amssymb,amsfonts}
\usepackage{algorithmic}
\usepackage{graphicx}
\usepackage{textcomp}
\usepackage{xcolor}
\usepackage{bbm,cite}
\usepackage{caption}
\usepackage{subcaption}

\usepackage[english]{babel}
\usepackage{amsthm}
\usepackage{url}

\theoremstyle{definition}
\newtheorem{definition}{Definition}

\begin{document}


\title{Admission Prediction in Undergraduate Applications: an Interpretable Deep Learning Approach \\
}

\author{\IEEEauthorblockN{Amisha Priyadarshini}
\IEEEauthorblockA{\textit{Department of Computer Science} \\
\textit{University of California, Irvine}\\
Irvine, CA, USA \\
apriyad1@uci.edu}

\and
\IEEEauthorblockN{ Barbara Martinez-Neda}
\IEEEauthorblockA{\textit{Department of Computer Science} \\
\textit{University of California, Irvine}\\
Irvine, CA, USA \\
barbarm@uci.edu}
\and

\IEEEauthorblockN{Sergio Gago-Masague}
\IEEEauthorblockA{\textit{Department of Computer Science} \\
\textit{University of California, Irvine}\\
Irvine, CA, USA \\
sgagomas@uci.edu}

}

\maketitle

\begin{abstract}
This article addresses the challenge of validating the admission committee's decisions for undergraduate admissions. In recent years, the traditional review process has struggled to handle the overwhelmingly large amount of applicants' data. Moreover, this traditional assessment often leads to human bias, which might result in discrimination among applicants. Although classical machine learning-based approaches exist that aim to verify the quantitative assessment made by the application reviewers, these methods lack scalability and suffer from performance issues when a large volume of data is in place. In this context, we propose deep learning-based classifiers, namely Feed-Forward and Input Convex neural networks, which overcome the challenges faced by the existing methods. Furthermore, we give additional insights into our model by incorporating an interpretability module, namely LIME. Our training and test datasets comprise applicants' data with a wide range of variables and information. Our models achieve higher accuracy compared to the best-performing traditional machine learning-based approach by a considerable margin of 3.03\%. Additionally, we show the sensitivity of different features and their relative impacts on the overall admission decision using the LIME technique.

\end{abstract}


\section{Introduction}
In recent decades, the undergraduate admission processes have witnessed significant transformations aimed at fostering fairness and equal opportunities for applicants. These changes are evident in the education system of the University of California (UC), which includes eliminating the consideration of race and gender, implementing the percentage plans to recognize the top-performing students, and removing the standardized testing along with the holistic review approach, which considers factors beyond the academic achievements of applicants. Integrating machine learning (ML) techniques in the undergraduate admissions decision-making process could ensure fairness by eliminating any human bias and prejudices that may inadvertently influence otherwise. Automating certain aspects of the decision process using ML models could also significantly enhance overall efficiency by saving time and resources for the admissions staff. \cite{Waters2014} and \cite{Gilbert2006} showcase two such software implementations of classical ML tools that could be used for the admissions process and analyze the hidden patterns in data. Another advantage of using ML classifiers is that they provide scalable solutions to tackle the increased workload by handling the increasing number of applications while ensuring every application's thorough and timely review. In support of our assertions, \cite{neda2022feasibility} explores the applicability and viability of classical ML models in the undergrad admissions process. The extensive simulation studies and findings thereof have led the way to multiple follow-up works like investigating the performance of state-of-the-art classifiers, such as Deep Neural Networks (DNN), for undergraduate admissions.

Deep learning (DL) has emerged as a powerful tool in various fields, from demonstrating their potential in understanding complex data to revolutionize the decision-making process. While classical ML models have shown promise, their performance in admission decisions may still be limited by their inability to capture intricate relationships within high-dimensional data. In contrast, DL models, with their multi-layer architecture, can learn hierarchical representations and are hence well suited for extracting meaningful information from complex and diverse applicant data. Furthermore, the flexibility and adaptability of deep neural networks(DNN) make them suitable for handling the dynamic nature of admissions data. These models can continuously learn and improve with new datasets, enabling universities to refine their decision-making processes.

Given the success of DNN models, the investigation of transparency, fairness and bias, concerning the domain of the admission decision process, still remains an active research area. To ensure the interpretability and explainability of the predictions obtained from the DNN models, we explore the Local Interpretable Model-agnostic Explanations (LIME) technique\cite{Ribeiro2016} coupled with a gradient-based approach. LIME aims to explain individual predictions made by complex models like DNNs. The gradient-based course considers the predicted output's sensitivity to any slight change in the input feature. We propose an approach where we mount the LIME technique together with the gradient-based feature selection to facilitate a generalized method. This could help extract invaluable information to understand an applicant's specific attributes and characteristics that heavily influenced admission decisions.

For the paper, we have considered 4,442 application records of California freshman applicants for the Fall 2021 cycle to the Department of Computer Science at UC, Irvine. The dataset encompassed a range of variables, including demographics, academic records, high school information, and essay question responses. Using the Python-based PyTorch framework, we trained the datasets on two different architectures of DNN models, namely Feed-Forward (FF) Neural Networks and Input Convex Neural Networks (ICNN). Additionally, we incorporated the LIME technique to offer interpretability and explanations for the predictions made by the DNN models. By presenting our findings, we aim to contribute to the existing body of knowledge on leveraging DNN models in undergraduate admission decisions while upholding the holistic review process.

\section{Related Works}
The use of ML models is gaining significant attention in recent years. However, a few studies have investigated the application of ML algorithms to predict and enhance the accuracy of admission decisions. Most of the studies conducted so far have focused on the graduate admission process, where the significant features influencing the decision process have been Undergrad Cumulative Grade Point Average (CGPA), Research, and Vacancies at research groups. In an undergraduate setting, neither these features hold any significance, nor could the results be generalized. Moreover, the majority of existing studies predominantly concentrate on numerical features instead of using the entire application and neglecting relevant aspects like text data. 
\subsection{Classical Machine Learning in Admission process}
For instance, \cite{Lux2016} has conducted a study for a specific University, where supervised learning techniques have been used to predict undergraduate admission decisions and the rate of student enrollment. The Support Vector Machine (SVM) has been reported to show high accuracy on training datasets consisting of academic information, family contribution, citizenship, and high school scores. Still, the study does not use text data nor offer any transparency of the predictions. Similarly, \cite{Andris2013} has focused on using supervised learning techniques with the SVM model reporting to obtain high accuracy in detecting undergraduate admission decision variation based on region. The work concentrates solely on standardized testing for training the models. However, according to \cite{sat2020}, various state universities have removed these standards owing to equity and access to preparation concerns.
\\In another line of work \cite{Waters2014}, depicts the development of a statistical machine learning system called GRADE assists in Ph.D. admissions at UT, Austin. It uses the historical admission data in predicting the admission decision and provides an insight into the feature significance using the Logistic regression model. Unfortunately, the model was discarded because of the non-updated training data and its counterproductive approach toward campus diversity \cite{Lilah2020}. Another such work \cite{Gilbert2006} suggests a software, Application Quest, which gives recommendations for graduate admission decisions that uphold diversity goals using the classical unsupervised clustering technique. The study reports the usage of the k-means algorithm to have the fastest execution time and is thus said to have achieved satisfactory results in university admission decision-making.
\\On the other hand, \cite{Staudaher2020} focuses on graduate applicant success in Georgia Tech's Online Masters in Analytics Program by using extracted features that quantify essays and Letters of Recommendation (LOR) instead of having the readers score them. This approach utilizes various supervised ML classifiers with Gradient Boosting reported to have provided the highest Area Under the Receiver Operating Characteristic Curve (AU-ROC) score. The article emphasizes preserving the holistic admission goals, and its favorable outcomes suggest potential in undergraduate admissions.

Many other papers focus on predicting student graduate admissions to help facilitate the process. For instance, \cite{sridhar2020university} has used the stacked ensemble models to predict graduate admissions accurately. Similarly, \cite{alghamdi2020machine} and \cite{sivasangari2021prediction} have also used classical ML models for predicting graduate and post-graduate admissions to help facilitate the university selection process for an aspiring student. Another study \cite{martinez2021using} talks regarding the bias in admission predictions, which we have tried to address in our study.

\begin{figure*} 
    \centering
  \subfloat[]{%
       \includegraphics[width=0.48\linewidth]{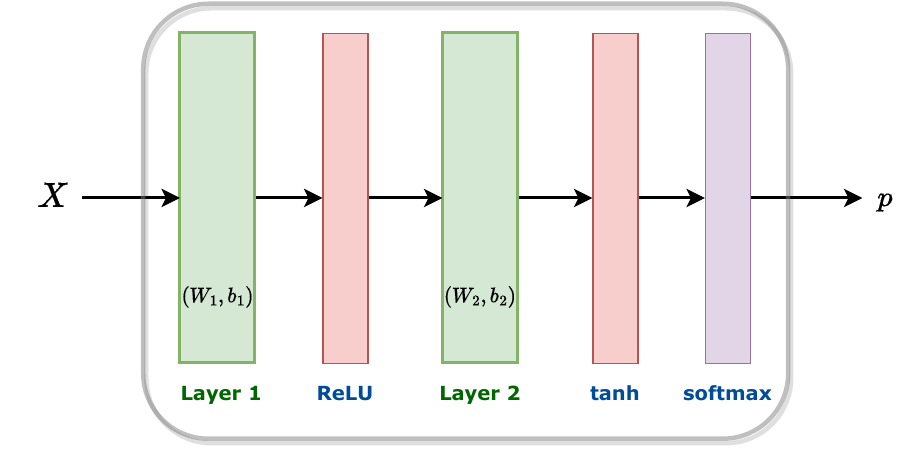}}
    \hfill
      \subfloat[]{%
       \includegraphics[width=0.42\linewidth]{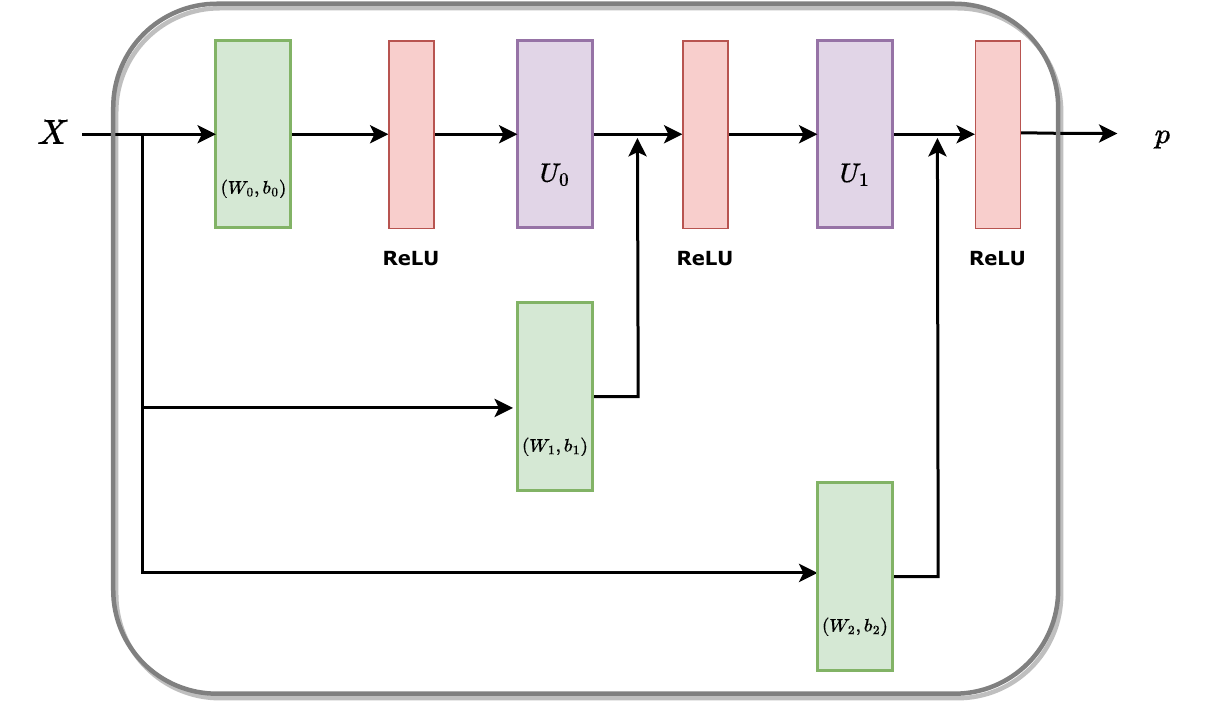}}
  \caption{Visual representation of deep neural network architectures, (a) Feed-Forward neural network; (b) Input convex neural network.}
  \label{fig:arch} 
\end{figure*}

\subsection{Deep Learning in Admission process}
 Within the framework of neural network applications, \cite{Alvero2020} emphasizes applicant essay analysis to understand and predict the demographic characteristics by training Logistic regression, Naive-Bayes, and Deep neural models on a corpus of application essays. The findings suggest Logistic regression is the most interpretable classifier, and infer data auditing to be useful in upholding holistic review. In another work \cite{Doleck2020}, the authors suggest comparing DL model predictions, using three different frameworks, with the classical ML techniques. The focus is on predicting the accuracy of student performance using MOOC (Massive open online course) and CEGEP (Collège d'enseignement général et professionnel) educational data sets. The study's findings indicate deep learning models achieve similar results as the classical ones, hence preferring the latter due to its higher interpretability. Both papers did not explore any interpretation techniques that could be deployed to understand the model's predictions and also, in turn, preserve transparency.
 \\Furthermore, \cite{Vasani2022} focuses on machine learning-based prediction for graduate admissions. The features were limited to English language proficiency scores and Research and Undergrad CGPA and are used to train 13 models categorized into deep learning, ensemble, and baseline models, respectively. Multiple-linear regression was reported to obtain the highest R2 score, but the number of data samples for training the models was scarce, making the predictions unreliable.

\section{Methodology}
This section provides an overview of the methodology, starting with data preprocessing, neural network training, and incorporating the LIME model for interpretability.
\subsection{Data preprocessing}\label{AA}
To facilitate the training of deep learning models and assess their performance in a comparison study with the classical machine learning models, we have obtained and pre-processed information on 4,442 applications to the CS department at UC, Irvine. The dataset used in this study comprises several categories of information, including the student Grade Point Average (GPA), Advanced Placement (AP) test scores, participation in educational programs, and responses to admission-related questions. Some dependent variables are demographic information, academic history, high school attended, and the responses to selected Personal Insight Questions (PIQs). PIQs offer insightful information regarding the students and their personalities, abilities, and experiences, which is valuable for model analysis\cite{piq}.

Given the problem we are trying to solve is a binary classification task, the final read score is assigned as the target variable, representing the review score assigned to every applicant. For this task, the top review score is mapped to 1, and the lower scores are mapped to 0. This approach seeks to identify the applications likely to receive high scores while ensuring a balanced dataset, with about 50\% records belonging to both classes.

The records missing a final read score value are excluded from the analysis to ensure the efficacy of the model training and performance evaluation. Next, to enhance the effectiveness of the model training process, any records with high school attendance outside California are excluded, considering that California applicants provide a wider range of potential information compared to out-of-state ones. Some of the necessary features present only on the California applicant records are data related to high school attendance, percentage value of the Eligibility in Local Context (ELC) program, etc. To meet the standards of holistic reviewing, it is of utmost importance to consider data that encompasses the individual accomplishments of the top 9 percent of students from each high school, considering the opportunities offered by their respective schools. Also, the records missing the numerical data, like the GPA scores, are dropped from the dataset, provided its high importance for the classification task. Following Proposition 209, we have dropped features like gender and ethnicity.

Features like primary major value consist of string entries, which are transformed into new binary columns after one-hot encoding, which indicated the presence or absence of the respective values from the original feature. Students were allowed to choose four PIQs to respond to from a collection of eight, which are later used to extract necessary information like character count, word count, sentence count, Flesh Reading Ease, Flesch-Kincaid grade level, polarity, subjectivity, and monosyllable vs. polysyllable word ratio. TextBlob\cite{textblob} and textstat\cite{textstat} libraries have been employed to extract the information from the PIQs. Provided, the Flesch Reading Ease formula is a widely accepted metric to assess text difficulty objectively\cite{Flesch1948}, which evaluates the readability and the comprehension difficulty of text-based syllables per 100 words and the average sentence length, with scores ranging from 0 to 100. Similarly, Flesch-Kincaid grade level is one of the metrics determining text difficulty, outputting a number corresponding to the grade level or relevant years of education\cite{Kincaid1975}. Then, the TextBlob library is used to perform sentiment analysis to extract the polarity and subjectivity values. Polarity ranges from -1 to 1 depending on the text's tone, and subjectivity ranges from 0 to 1, with 0 representing more objective text and 1 more subjective text\cite{ttr}. After extracting this information, the text columns have been dropped from the dataset.

After obtaining a tabular dataset, we handle the missing entries by the method of median imputation, where the missing values are replaced with the corresponding feature's median value. This ensures robustness to outliers or extreme values and helps preserve the variable's overall distribution. This step is followed by data normalization, which transforms the data to have zero mean and unit variance. It, in turn, prevents feature dominance and improves performance by ensuring data convergence for gradient-based optimizations performed in a neural network. Subsequently, data scaling was performed to map the normalized data to the range of [0, 1], which is beneficial for algorithms relying on specific input ranges, like the DNN models.

\subsection{Neural Network Training}
Neural networks have emerged as a powerful and versatile tool in the field of ML, which has been designed to mimic the functioning of the human brain loosely. They hold the capability of handling and mapping complex relationships within the data. The structure of a neural network typically consists of an input layer, one or more hidden layers, and an output layer.


In the paper, we focus on two different neural network architectures, namely Feed-Forward (FF) and Input convex neural networks (ICNN), for predicting undergraduate admission decisions. We have discussed hyper-parameters and their mathematical significance for the training process. The implemented setup for the various hyper-parameters used in the study has been demonstrated in Table \ref{tab1}.

\begin{table}[t]
\caption{Hyper-Parameters of the Proposed Models}
\begin{center}
\renewcommand{\arraystretch}{1.2}
\begin{tabular}{l l}
\hline
\textbf{Feed-Forward Neural Network} \\
\hline
Data Loader batch size & 64\\
Neurons in the first hidden layer & 256\\
Activation function for first hidden layer & ReLu\\
Neurons in the second hidden layer & 32\\
Activation function for the second hidden layer & tanh\\
Optimizer & ADAM\\
Learning rate & 0.001\\
Loss function & Cross-Entropy\\
L2-regularization term, $\lambda$ & 0.01\\

\hline

\textbf{Input Convex Neural Network} \\
\hline
Data Loader batch size & 64\\
Neurons in the first hidden layer & 512\\
Activation function for first hidden layer & ReLu\\
Neurons in the second hidden layer & 16\\
Activation function for the second hidden layer & ReLu\\
Optimizer & ADAM\\
Learning rate & 0.001\\
Loss function & Cross-Entropy\\
L2-regularization term, $\lambda$ & 0.001\\
Dropout layer Probability & 0.3\\

\hline
\end{tabular}
\label{tab1}
\end{center}
\end{table}

\subsubsection{Feed-Forward Neural Network}
The Feed-Forward neural network is a fundamental architecture where the information propagates within the network in a unidirectional manner, originating from the input layer and progressing toward the output layer. The network lacks any recurrent or feedback loops. We consider a three-layered architecture or a deep neural network with two hidden layers for the binary classification problem. The proposed FF architecture is illustrated in Figure \ref{fig:arch}(a).
For a feed-forward neural network, its input is a state vector and its output is a scalar value. We write $x$ as the input to the network with a bias $b$ and weight vector $W$. For notational convenience, we consider the activation functions as mentioned in Table \ref{tab1}.

\begin{definition}[Feed-Forward Predicted Output]
    The predicted output of the network, $\text{p}$, is defined as the following,
    \[p = \text{softmax}[~ \text{tanh}(W_2 . ~\text{ReLU}(W_1 . x + b_1) + b_2) ]\]
    where $W_1$ and $W_2$ are the weight vectors and, $b_1$ and $b_2$ are the bias vectors respectively. 
\end{definition}
We have used $\text{tanh}$ and $\text{ReLU}$ activation functions on the first and second hidden layers, respectively. Despite using the softmax activation function on the output layer of a binary classification problem, it has shown a tremendous improvement in the model's performance along with the cross-entropy loss function. 

This architecture enables the network to learn the hierarchical representations of the input data. We employ the ADAM\cite{kingma2014adam} optimizer which is a first-order gradient-based optimization technique to update the weights and bias based on the calculated gradients of the loss function accordingly.

\subsubsection{Input Convex Neural Network}
The second type of architecture used for the study is the ICNN, a scalar-valued neural network with constraints on its parameters or weights such that the output of the network is a convex function of the inputs\cite{amos2017input}. The architecture ensures convexity, a fundamental concept in the optimization theory that allows efficient and reliable optimization, explicitly capturing the convex relationships between the input variables. For the study, we have considered a fully connected Input Convex neural network (FICNN) consisting of several passthrough layers. The significance of including passthrough layers between the hidden layers can be beneficial to preserving the convexity of the input space while allowing the neural network to perform non-linear transformations to understand the complex patterns in data. An illustration of the said model is provided in Figure \ref{fig:arch}(b).
\begin{definition}[FICNN Predicted Output]
    Say, $f(x;\theta)$ is the scalar-valued neural network where $x$ denotes the input to the function and $\theta$ are the parameters such that the network, $f$, is convex to the input $x$.
    In order to understand the predictions, the layer-wise output is calculated as,
    \[z_{i+1} = g_i( U_i.z_i + W_i.x + b_i) \]
    where $W_i$ are the real-valued weights mapping from inputs to the $i+1$ layer activations; $U_i$ are positive weights mapping previous layer activations $z_i$ to the next layer; $b_i$ are the real-valued bias terms; and $g_i$ are convex, monotonically non-decreasing non-linear activation functions for every $i$ representing the training samples.
    Then the predicted output of the network could be defined as follows,
    \[p = g(x) \equiv z_k \]
    where k signifies the number of layers in the network.
\end{definition}


Furthermore, we incorporate the Dropout regularization technique in the architecture, enhancing the model generalization and robustness to noise and variation in the input data, simultaneously reducing model sensitivity. The network imposes constraints to preserve the convexity with respect to the input, which ensures the output has convex regions in the input space. The significance of the model lies in its ability to preserve convexity leading to more reliable solutions, and helps facilitate optimization to yield higher performance metrics.

\subsubsection{Principal Component Analysis}
In the next step, we incorporate the Principal Component Analysis (PCA) with both the neural network architectures discussed to check for any significant improvements in the model prediction. PCA is a dimensionality reduction technique implemented to transform a high-dimensional dataset into a low-dimensional one while preserving the important features. The technique is applied to address the challenges associated with high-dimensional datasets, potentially improving the model's performance and robustness to noise and redundancy.

\begin{table*}[htp]
\caption{Performance metrics for deep neural network models}
\begin{center}
\renewcommand{\arraystretch}{1.3}
\begin{tabular}{c c c c c c}
\hline
\textbf{Neural Network Model Architecture} & \textbf{\textit{Accuracy}}& \textbf{\textit{Precision}}& \textbf{\textit{Recall}}& \textbf{\textit{F1-Score}}& \textbf{\textit{AU-ROC Score}} \\
\hline
Feed-Forward Neural Network & 0.8056 & 0.8159 & 0.7883 & 0.8018 & 0.8056 \\
Feed-Forward Neural Network with PCA & \textbf{0.8067} & 0.8000 & \textbf{0.8073} & 0.8037 & 0.8068 \\
ICNN & \textbf{0.8067} & \textbf{0.8281} & 0.7961 & \textbf{0.8118} & \textbf{0.8073} \\
ICNN with PCA & 0.8056 & 0.8248 & 0.7983 & 0.8113 & 0.8060 \\
\hline
\end{tabular}
\label{tab2}
\end{center}
\end{table*}

\subsection{Prediction Interpretability using LIME}
Local Interpretable Model-agnostic Explanations (LIME) is a technique used to interpret the predictions of complex machine learning models like deep neural networks. It is particularly used when the ML model's inner workings are opaque or difficult to interpret. Being a model-agnostic method, it tries to learn the underlying behavior of the ML model by perturbing the input and observing the changes in the model predictions. As the LIME model is only capable of providing local explanations by approximating the behavior of the ML model in the vicinity of a data instance, we have tried to adopt a slightly different technique to improve the transparency of model predictions by generalizing them over a broader spectrum of testing samples. To cascade over the limitation of the LIME model, we have tried to do the feature selection using the Gradients in the FF neural network architecture, followed by applying the LIME model on the neural network trained over the selected features. The reason behind the choice of architecture is that we wanted the model to be comparatively easy to interpret and thus help us in understanding the predictions more efficiently. We try to compute the gradients of the neural network with respect to the input features, which is the measure of the sensitivity of the output due to any perturbation in the input.

\begin{definition}[Gradient Computation]
    For a fully connected neural network with cross-entropy loss function, $L$, we compute the gradients as follows
    \[\nabla_{\theta}L = \delta{L}/\delta{\theta}
    \]
    \[\theta = \{ W_i, b_i\}\]
    where $W_i$ is the weight, $b_i$ is the bias of $i$-th layer of the neural network.

\end{definition}

Next, we implement the LIME model on the neural network trained on the selected features. This step ensures the dimensionality reduction necessary to eliminate redundant features for better performance and improve the scope of interpretability. In order to tackle the local explanations provided by LIME, we intend to combine the gradient-based feature selection to improve the results by evaluating the model's behavior on a diverse range of instances and perturbing a selected set of features. This helps in understanding the patterns and behaviors beyond individual predictions, providing an overview of the model's decision-making process and in turn an attempt at deriving a global explanation using the LIME model.
The local explanations provided by the LIME model help in understanding the model's decisions on specific instances and the global explanations provided through the above process help reveal the pattern and model's feature selection trend across the dataset.

\section{Simulation Results}
In this section, we discuss the various performance metrics of our DNN model for student admissions and provide a comparative study with the classical ML models. We further discuss the two DL models used and the significance of feature importance in the admissions process.
\subsection{Performance Metrics}
We have implemented and trained the FF and the ICNN, deep learning models on the dataset with 80\% used for training and 20\% for testing. We have utilized Python-based PyTorch\cite{paszke2019pytorch} as the DL framework and trained the DL models using the standard backpropagation procedure and included the related code\footnote{The GitHub repository link can be found at https://github.com/apriyad1/Deep-Learning-in-Admissions.}. To evaluate the model's performance we have employed five types of metrics, namely accuracy, precision, recall, F1-score, and AU-ROC score. As opposed to the classical ML performance in \cite{neda2022feasibility} we have successfully been able to showcase significant improvements in performance using the same dataset for the student admission problem in our paper. 
The accuracy metric is chosen as one of the performance metrics obtained by dividing the number of correct predictions by the total number of records in the dataset, providing a percentage of correct predictions. Owing to the balanced target class, the accuracy metric makes for a suitable and reliable measure to assess model performance. Apart from that the F1-score and the recall metrics served as effective metrics for understanding the DL model performance. The F1-score represents the harmonic mean of precision and recall, and the recall represents the proportion of true positive predictions out of all actual positive instances.

The ICNN and the FF models both seem to achieve an accuracy of 0.8067, outperforming the classical ML models by a substantial margin. We have also been able to outperform the classical ML models by demonstrating higher precision and F1 scores. As observed from Table \ref{tab2}, the ICNN model outperforms the FF model on four metrics including precision, recall, F1-score, and AUROC score, with Accuracy remaining the same as the FF model employed with PCA. Based on the overall statistics, the ICNN model shows promising results in solving the student admission decision problem. The superior performance obtained by the ICNN model can be attributed to its intrinsic convexity that results in robust and efficient non-linear decision boundaries in input space. We note that our simulation results for ICNN conform to the findings of \cite{amos2017input} concerning classification tasks.

\begin{figure} 
    \centering
  \subfloat[]{%
       \includegraphics[width=0.6\linewidth]{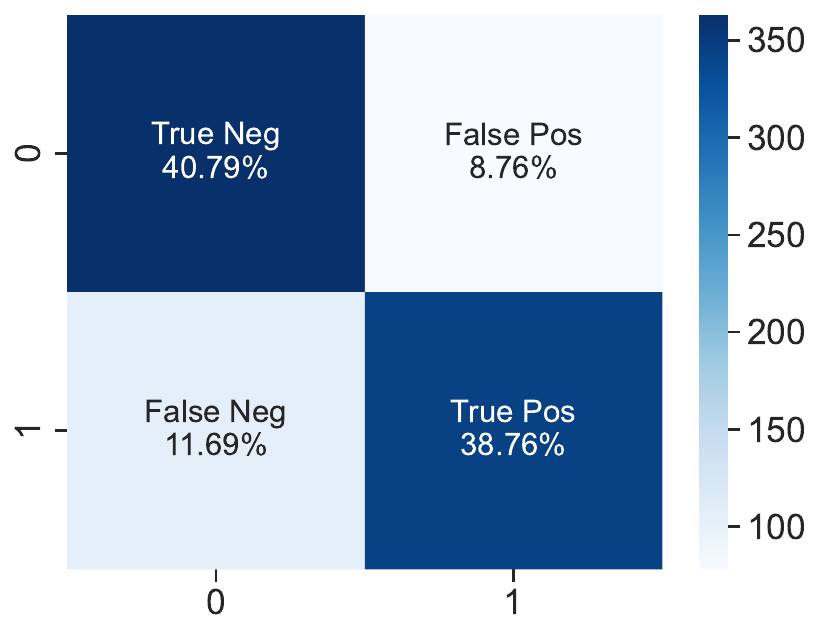}}
    \vfill
      \subfloat[]{%
       \includegraphics[width=0.6\linewidth]{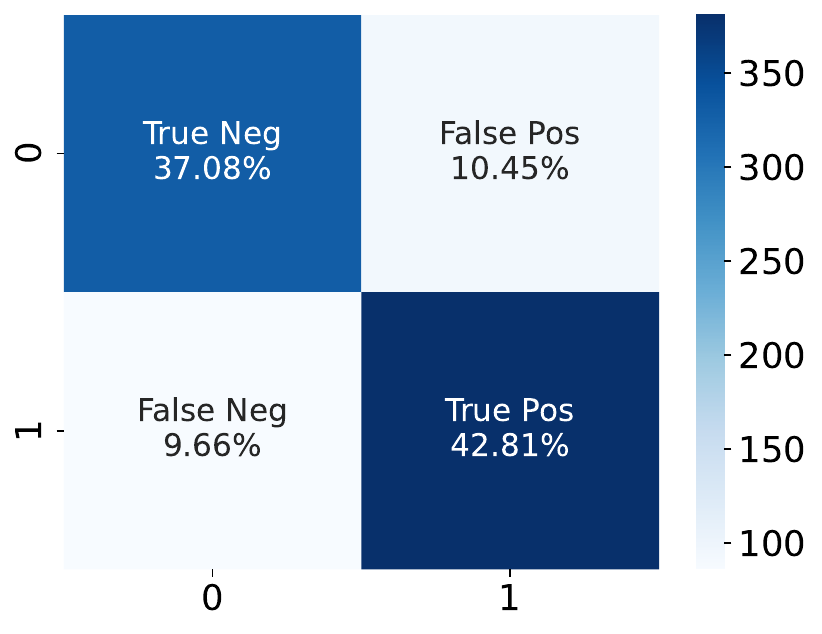}}
  \caption{Confusion matrices for fully-trained deep learning models, (a) Feed-Forward neural network; (b) Input convex neural network.}
  \label{fig:conf} 
\end{figure}
 
In Figure \ref{fig:conf} we present the confusion matrices for both the FF neural network and ICNN models. The confusion matrix has been provided to give a visual representation of the models' relative performance on the test dataset in predicting admission decisions and thereby evaluating their metrics.
As discussed previously, we have incorporated the LIME model using the gradient-based approach for understanding the feature significance of the Deep neural network model. This augmentation leverages the gradients of the Feed-forward neural network model's output with respect to the input features to assess their contribution to the predicted output. We then employ the LIME model to further understand and differentiate the impact of each selected feature on the test dataset.

\subsection{Discussion}

\begin{figure*}
\centering
\includegraphics[width=0.8\linewidth]{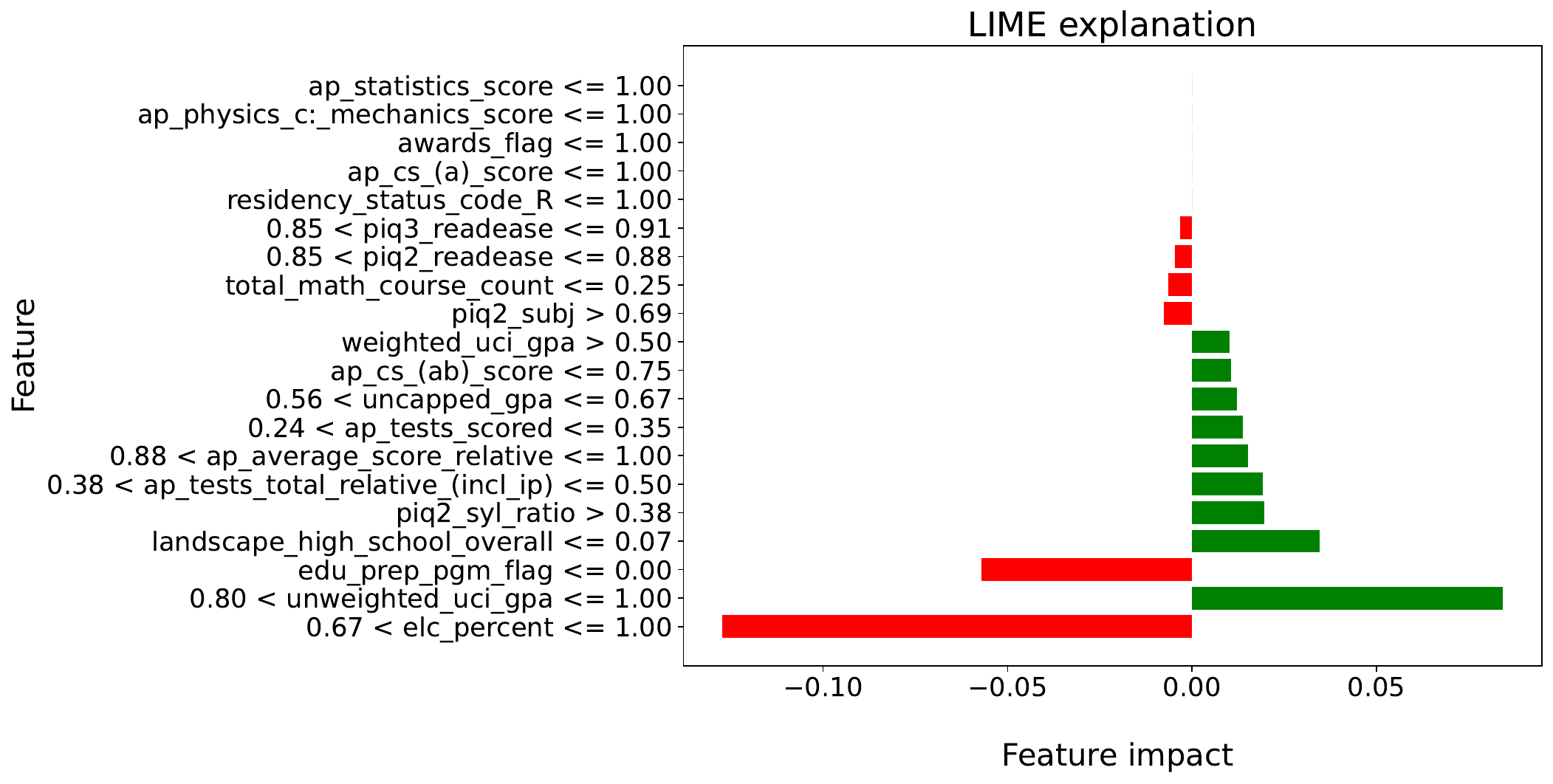}
\caption{LIME interpretation of the Feed-Forward model predictions. The green and red horizontal bars signify the key features influencing  the overall classification in positive and negative ways respectively.}
\label{fig:lime}
\end{figure*}

In this paper, we discuss the usage of a gradient-based method to initially select the top 20 highly affecting features from our trained Feed-forward neural network model for the student admissions problem. Subsequently, applying the LIME  technique to the selected features, we try to analyze the positive and negative feature impacts on our decision-making process. Every class of feature is assigned a probability score based on the possibility of getting chosen. This method enhances our chances to understand the feature significance in the decision-making process over a broader aspect and hence tackle the local interpretability.

As illustrated in Figure \ref{fig:lime}, the features signifying academic performance, including GPA, namely the Unweighted and Weighted GPAs, and performance in AP tests such as the AP tests total(relative) scores, average AP scores, and AP CS A score, act as the major key features in influencing the admission decision process in a positive manner, which has been marked in green. This suggests that students with higher academic achievement are more likely to be admitted as opposed to the ones with lesser achievement. The performance in the AP tests, as shown to influence positively, showcases student competence in challenging coursework and also reflects student's ability to handle rigorous academic pressure. Similarly, the high school Landscape score also accounts for heavily affecting the decision process in a positive manner. This feature helps the admission decisions to be informed and equitable by considering an applicant's accomplishments within their high school environment mitigating any kind of unfair bias. Moreover, the PIQ2 Syllable ratio also accounts for positively affecting admission decisions, providing insights into the applicant's communication skills, critical thinking ability, and writing.

At the same time, certain features are found to impact the student admission decision process negatively, which has been marked in red. Out of all, the ELC percentage score and the Education Preparation Programs(EPP) involvement flag has the highest negative influence on the decision-making process. 
The ELC score recognizes the achievements of the top 9\% of students from each high school in California. For example, students with a value of 1 are the top 1\% performers in their high school, while a 9 means that students are in the top 9\% of their class. Those who are not in the top 9\% did not receive an ELC score and their records were imputed with a value of 10. As a result, the negative LIME score associated with the ELC feature shows that students who had a higher value, meaning that they were further away from the top-performing status, were less likely to be classified as admitted. This ensures a fair evaluation, considering the opportunities and challenges faced by the applicant, and promoting equity in the admissions process.

Continuing on, the EPP flag indicates whether the students participated in any kind of educational preparation programs or not. Given our analysis, students who participated in these programs were less likely to be admitted. Apart from that, the Total Maths score count, PIQ2 and PIQ3 Flesch-reading scores also have a negative influence on the decision-making process. Additionally, we have identified a set of features that demonstrate a negligible effect on admission decisions, including the  AP statistics scores, the AP Physics C score, and the awards flag. While these features did not exert significant influence individually, they still highly contribute to the overall assessment of an applicant.

Furthermore, the results obtained using the LIME interpretability module in our study have been found to be mostly consistent with the feature coefficients (another way to interpret classical ML models) derived in the preliminary paper\cite{neda2022feasibility}. It further validates the influence and importance of the identified features, reinforcing our model to be robust and reliable in capturing the key factors contributing to the applicant evaluation for the undergraduate admissions process. We note that by highlighting the overall impact of the features, our study emphasizes on a broad spectrum including academic aptitude and holistic qualities while evaluating the applicants for undergraduate student admissions.

\section{Conclusion and Future Works}
In this paper, we have demonstrated the superior performance showcased by the Deep neural network models  in the undergraduate student admission decision-making process. The ICNN model outperforms the remaining baselines by a considerable margin, including the classical models. We have achieved high accuracy in our experiments with both the ICNN and FF neural network models making them a suitable choice for the task, further affirming their effectiveness.
Furthermore, we have leveraged a gradient-based course coupled with the LIME model to extract significant features responsible for the decision process. The approach allowed us to understand the feature importance and the various types of impact it has on the admission process. Through our analysis, we have identified certain key features that positively influence the decision-making process, including the GPA-based features and some selected AP scores which highlight the importance of academic achievements, educational backgrounds, and performance in advanced coursework in shaping admission outcomes. Additionally, we have also identified negatively affecting features, like the ELC percent and EPP flag, which furthers our need for a holistic evaluation procedure that takes into consideration various factors beyond academic performance. In summary, the application of the ML prediction interpretation technique provides valuable insights which further enable a deeper understanding of the feature significance and contribute to enhancing transparency, fairness, and accountability in the admission process.
\\In this study, we have made significant strides in understanding the undergraduate student admissions process using deep neural network models and prediction interpretability to understand them further. However, there are several potential future research directions that could promote transparency and ensure fairness, and therefore help in upholding the holistic review. One such promising direction could be to explore anomaly detection in order to identify unusual patterns or outliers in the admission data. By flagging suspicious cases, institutions could conduct further investigations and validations to assure that future admission decisions are more reliable. Furthermore, exploring model-agnostic techniques for anomaly detection in the admission process could improve robustness by detecting anomalies in any type of admissions model. This future research holds the potential to contribute towards developing more robust and qualified admission systems and thereby mitigating any bias or inconsistency in the decision-making process.






\bibliographystyle{IEEEtran}
\bibliography{main}

\begin{thebibliography}{10}
\providecommand{\url}[1]{#1}
\csname url@samestyle\endcsname
\providecommand{\newblock}{\relax}
\providecommand{\bibinfo}[2]{#2}
\providecommand{\BIBentrySTDinterwordspacing}{\spaceskip=0pt\relax}
\providecommand{\BIBentryALTinterwordstretchfactor}{4}
\providecommand{\BIBentryALTinterwordspacing}{\spaceskip=\fontdimen2\font plus
\BIBentryALTinterwordstretchfactor\fontdimen3\font minus
  \fontdimen4\font\relax}
\providecommand{\BIBforeignlanguage}[2]{{%
\expandafter\ifx\csname l@#1\endcsname\relax
\typeout{** WARNING: IEEEtran.bst: No hyphenation pattern has been}%
\typeout{** loaded for the language `#1'. Using the pattern for}%
\typeout{** the default language instead.}%
\else
\language=\csname l@#1\endcsname
\fi
#2}}
\providecommand{\BIBdecl}{\relax}
\BIBdecl

\bibitem{Waters2014}
\BIBentryALTinterwordspacing
A.~Waters and R.~Miikkulainen, ``{Grade: Machine-learning support for graduate
  admissions},'' in \emph{AI Magazine}, vol.~35, no.~1, 2014, pp. 64--75.
  [Online]. Available: \url{www.aaai.org}
\BIBentrySTDinterwordspacing

\bibitem{Gilbert2006}
J.~E. Gilbert, ``{Applications Quest: Computing Diversity},'' Auburn
  University, Tech. Rep., 2006.

\bibitem{neda2022feasibility}
Anonymous, ``Anonymous,'' in \emph{Anonymous}.\hskip 1em plus 0.5em minus
  0.4em\relax IEEE, 2022, p. Anonymous.

\bibitem{Ribeiro2016}
\BIBentryALTinterwordspacing
M.~T. Ribeiro, S.~Singh, and C.~Guestrin, ``{"Why should i trust you?"
  Explaining the predictions of any classifier},'' in \emph{Proceedings of the
  ACM SIGKDD International Conference on Knowledge Discovery and Data Mining},
  2016, pp. 1135--1144. [Online]. Available:
  \url{http://dx.doi.org/10.1145/2939672.2939778}
\BIBentrySTDinterwordspacing

\bibitem{Lux2016}
\BIBentryALTinterwordspacing
T.~Lux, R.~Pittman, M.~Shende, and A.~Shende, ``{Applications of supervised
  learning techniques on undergraduate admissions data},'' \emph{2016 ACM
  International Conference on Computing Frontiers - Proceedings}, pp. 412--417,
  may 2016. [Online]. Available:
  \url{http://dx.doi.org/10.1145/2903150.2911717}
\BIBentrySTDinterwordspacing

\bibitem{Andris2013}
C.~Andris, D.~Cowen, and J.~Wittenbach, ``{Support Vector Machine for Spatial
  Variation},'' \emph{Transactions in GIS}, vol.~17, no.~1, pp. 41--61, feb
  2013.

\bibitem{sat2020}
\BIBentryALTinterwordspacing
``University of california will no longer consider sat and act scores.''
  [Online]. Available:
  \url{https://www.nytimes.com/2021/05/15/us/SAT-scores-uc-university-of-california.html}
\BIBentrySTDinterwordspacing

\bibitem{Lilah2020}
\BIBentryALTinterwordspacing
``The death and life of an admissions algorithm.'' [Online]. Available:
  \url{https://www.insidehighered.com/admissions/article/2020/12/14/u-texas-will-stop-using-controversial-algorithm-evaluate-phd}
\BIBentrySTDinterwordspacing

\bibitem{Staudaher2020}
\BIBentryALTinterwordspacing
S.~Staudaher, J.~Lee, and F.~Soleimani, ``{Predicting Applicant Admission
  Status for Georgia Tech's Online Master's in Analytics Program},'' \emph{L@S
  2020 - Proceedings of the 7th ACM Conference on Learning @ Scale}, pp.
  309--312, aug 2020. [Online]. Available:
  \url{http://dx.doi.org/10.1145/3386527.3406735}
\BIBentrySTDinterwordspacing

\bibitem{sridhar2020university}
S.~Sridhar, S.~Mootha, and S.~Kolagati, ``A university admission prediction
  system using stacked ensemble learning,'' in \emph{2020 Advanced Computing
  and Communication Technologies for High Performance Applications
  (ACCTHPA)}.\hskip 1em plus 0.5em minus 0.4em\relax IEEE, 2020, pp. 162--167.

\bibitem{alghamdi2020machine}
A.~AlGhamdi, A.~Barsheed, H.~AlMshjary, and H.~AlGhamdi, ``A machine learning
  approach for graduate admission prediction,'' in \emph{Proceedings of the
  2020 2nd International Conference on Image, Video and Signal Processing},
  2020, pp. 155--158.

\bibitem{sivasangari2021prediction}
A.~Sivasangari, V.~Shivani, Y.~Bindhu, D.~Deepa, and R.~Vignesh, ``Prediction
  probability of getting an admission into a university using machine
  learning,'' in \emph{2021 5th International Conference on Computing
  Methodologies and Communication (ICCMC)}.\hskip 1em plus 0.5em minus
  0.4em\relax IEEE, 2021, pp. 1706--1709.

\bibitem{martinez2021using}
B.~Martinez~Neda, Y.~Zeng, and S.~Gago-Masague, ``Using machine learning in
  admissions: Reducing human and algorithmic bias in the selection process,''
  in \emph{Proceedings of the 52nd ACM Technical Symposium on Computer Science
  Education}, 2021, pp. 1323--1323.

\bibitem{Alvero2020}
\BIBentryALTinterwordspacing
A.~J. Alvero, N.~Arthurs, A.~L. Antonio, B.~W. Domingue, B.~Gebre-Medhin,
  S.~Giebel, and M.~L. Stevens, ``{AI and holistic review: Informing human
  reading in college admissions},'' in \emph{AIES 2020 - Proceedings of the
  AAAI/ACM Conference on AI, Ethics, and Society}, 2020, pp. 200--206.
  [Online]. Available: \url{https://doi.org/10.1145/3375627.3375871}
\BIBentrySTDinterwordspacing

\bibitem{Doleck2020}
\BIBentryALTinterwordspacing
T.~Doleck, D.~J. Lemay, R.~B. Basnet, and P.~Bazelais, ``Predictive analytics
  in education: A comparison of deep learning frameworks,'' \emph{Education and
  Information Technologies}, vol.~25, no.~3, pp. 1951--1963, may 2020.
  [Online]. Available: \url{https://doi.org/10.1007/s10639-019-10068-4}
\BIBentrySTDinterwordspacing

\bibitem{Vasani2022}
M.~Vasani, S.~Patel, and J.~Kaur, ``Comparative analysis of baseline models,
  ensemble models, and deep models for prediction of graduate admission,'' in
  \emph{Proceedings of 3rd International Conference on Machine Learning,
  Advances in Computing, Renewable Energy and Communication: MARC 2021}.\hskip
  1em plus 0.5em minus 0.4em\relax Springer, 2022, pp. 515--525.

\bibitem{piq}
``Personal insight questions,''
  \url{https://admission.universityofcalifornia.edu
  /how-to-apply/applying-as-a-freshman/personal-insight-questions.html}.

\bibitem{textblob}
\BIBentryALTinterwordspacing
``Simplified text processing.'' [Online]. Available:
  \url{https://textblob.readthedocs.io/en/dev/}
\BIBentrySTDinterwordspacing

\bibitem{textstat}
\BIBentryALTinterwordspacing
``Textstat.'' [Online]. Available:
  \url{https://textstat.readthedocs.io/en/latest/}
\BIBentrySTDinterwordspacing

\bibitem{Flesch1948}
R.~Flesch, ``{A new readability yardstick},'' \emph{Journal of Applied
  Psychology}, vol.~32, no.~3, pp. 221--233, 1948.

\bibitem{Kincaid1975}
\BIBentryALTinterwordspacing
J.~Kincaid, R.~{Fishburne Jr.}, R.~Rogers, and B.~Chissom, \emph{Naval
  Technical Training Command Millington TN Research Branch}, 1975. [Online].
  Available:
  \url{http://oai.dtic.mil/oai/oai?verb=getRecord&metadataPrefix=html&identifier=ADA006655}
\BIBentrySTDinterwordspacing

\bibitem{ttr}
\BIBentryALTinterwordspacing
``Simplified text processing.'' [Online]. Available:
  \url{https://carla.umn.edu/learnerlanguage/spn/comp/activity4.html}
\BIBentrySTDinterwordspacing

\bibitem{kingma2014adam}
D.~P. Kingma and J.~Ba, ``Adam: A method for stochastic optimization,''
  \emph{arXiv preprint arXiv:1412.6980}, 2014.

\bibitem{amos2017input}
B.~Amos, L.~Xu, and J.~Z. Kolter, ``Input convex neural networks,'' in
  \emph{International Conference on Machine Learning}.\hskip 1em plus 0.5em
  minus 0.4em\relax PMLR, 2017, pp. 146--155.

\bibitem{paszke2019pytorch}
A.~Paszke, S.~Gross, F.~Massa, A.~Lerer, J.~Bradbury, G.~Chanan, T.~Killeen,
  Z.~Lin, N.~Gimelshein, L.~Antiga \emph{et~al.}, ``Pytorch: An imperative
  style, high-performance deep learning library,'' \emph{Advances in neural
  information processing systems}, vol.~32, 2019.

\end{thebibliography}

\end{document}